\documentclass[journal]{IEEEtran}
\usepackage{cite}

\ifCLASSINFOpdf
   \usepackage[pdftex]{graphicx}
\else
\fi
\usepackage{amsmath}
\usepackage{amsfonts}
\usepackage{multirow}
\usepackage{float}
\usepackage{bbm}
\usepackage{nicefrac}
\usepackage[ruled,vlined]{algorithm2e}

\usepackage{setspace}

\usepackage{soul} 
\usepackage{booktabs} 

\usepackage[colorlinks=true,linkcolor=blue,citecolor=blue,urlcolor=black ]{hyperref}

\usepackage{graphicx}

\usepackage{subcaption}

\begin{document}
%
\title{DeepWay: a Deep Learning Waypoint Estimator for Global Path Generation}
%
%
%

\author{Vittorio~Mazzia, Francesco~Salvetti, Diego~Aghi, and~Marcello~Chiaberge
\thanks{The authors are with Politecnico di Torino -- Department of Electronics and Telecommunications, PIC4SeR, Politecnico di Torino Interdepartmental Centre for Service Robotics and SmartData@PoliTo, Big Data and Data Science Laboratory, Italy. Email: \{name.surname\}@polito.it.}}

%
%

\markboth{}%
{Vittorio Mazzia \MakeLowercase{\textit{et al.}}: DeepWay: a Deep Learning Waypoint Estimator for Global Path Generation}
%



\maketitle

\begin{abstract}
Agriculture 3.0 and 4.0 have gradually introduced service robotics and automation into several agricultural processes, mostly improving crops quality and seasonal yield. Row-based crops are the perfect settings to test and deploy smart machines capable of monitoring and manage the harvest. In this context, global path generation is essential either for ground or aerial vehicles, and it is the starting point for every type of mission plan. Nevertheless, little attention has been currently given to this problem by the research community and global path generation automation is still far to be solved. 
In order to generate a viable path for an autonomous machine, the presented research proposes a feature learning fully convolutional model capable of estimating waypoints given an occupancy grid map. In particular, we apply the proposed data-driven methodology to the specific case of row-based crops with the general objective to generate a global path able to cover the extension of the crop completely. Extensive experimentation with a custom made synthetic dataset and real satellite-derived images of different scenarios have proved the effectiveness of our methodology and demonstrated the feasibility of an end-to-end and completely autonomous global path planner.
\end{abstract}


%

\section{Introduction}  

Over the past years, several research activities related to precision agriculture and smart farming have been published \cite{kamilaris2017review,kamilaris2018deep,jhuria2013image,tripicchio2015towards, mazzia2020real}, sign of a new industrial revolution approaching the agricultural world. Agriculture 4.0 brought a new concept of agriculture based on the introduction of robotics, artificial intelligence and automation into the agricultural processes in order to increase production efficiency and to cut labour costs. 
In this regard, self-driving agricultural machinery plays a relevant role both in production efficiency, by providing a 24/7 weather-independent working production system, and cost-cutting, since there is not the need of a paid driver when performing the required task anymore. Moreover, these systems can be used as a support for an autonomous harvesting \cite{kang2020real,luo2018vision} or for plant and fruit disease detection \cite{mohanty2016using,ferentinos2018deep,khaliq2019refining,mazzia2020uav}.
Indeed, in a row-based crop's environment, many works on autonomous navigation systems have been carried out by using deep learning and computer vision techniques \cite{aghi2020autonomous,aghi2020local} or with sensor-based approaches  \cite{riggio2018low,astolfi2018vineyard,barawid2007development, mazzia2019deep}.

A good path generator is crucial for obtaining high autonomous navigation performance. However, in this type of lands, the global path generation automation problem has been a bit neglected by the research community. 
Nevertheless, the most common solutions for this task are based on clustering techniques applied on satellite images or aerial footage taken from the drones. For instance, in \cite{zoto2019automatic}, authors use clustering in order to detect the rows of the vineyards from UAV images, and then the trajectory is computed by exploiting the information given by the clusters. 
As shown in \cite{vidovic2014center}, extrapolating information regarding the row crops from the images is complex and computational heavy, and even though there are other solutions besides clustering such as \cite{comba2018unsupervised}, the complete pipeline for obtaining a global path is still tricky and time consuming due to this necessity of information regarding the crops position and orientation. 

In this regard, we introduce DeepWay, a novel deep learning approach for global path planning generation of row-based crop environments. As input, it requires just an occupancy grid of the analyzed parcel and provides, as output, a trajectory able to cover each row of the considered crop avoiding unwanted collisions with fixed obstacles.
The deep neural network is trained on a carefully devised synthetic dataset and is designed to predict global path waypoints directly from the binary mask of crops. Successively, output waypoints are processed with a  refinement pipeline in order to remove spare waypoints and to add missing ones. Finally, the global path is computed through the A* search algorithm. Extensive experimentation with the synthetic dataset and real satellite-derived images of different scenarios are used to validate the proposed methodology.
All of our training and testing code and data are open source and publicly available \footnote{https://github.com/fsalv/DeepWay}.

The rest of the paper is organized as follows. Section 2 covers the synthetic dataset design and generation. In section 3, the proposed methodology is analyzed with a detailed explanation of the DeepWay architecture and the waypoint refinement and path generation processes. Finally, section 4 presents the experimental results and discussion followed by the conclusion.

\section{Dataset Construction}
Due to the lack of online datasets of row crop occupancy grids and the complexity of building a real one in scale, we carefully devise a synthetic one. We mainly focus on linear crops, since the majority of the real-world case enters in this category.Thanks to the geometrical simplicity of a row crop field, we design an algorithm to generate any number of occupancy grids of shape $H \times W$ with a random number of rows $N$ and angle $\alpha$. We select $N=20$ as the minimum number of rows per image, and 50 as maximum. $\alpha$ can be any angle between $-\pi/2$ and $\pi/2$. The images are generated as single-channel masks with 1-bit values: 0 for the background and 1 for the occupied pixels.

$N$ points are identified as centres of the rows along a line perpendicular to $\alpha$ starting from the image centre. To take in consideration any possible orientation of the field with respect to the point of view and any possible angle between the rows and the field edges, we generate borders with randomly orientations and we define the first and the last point of each row such that the line that connects them passes trough the row centre and has an orientation equal to $\alpha$. To further increase the variability of generated images, a random displacement is added to the coordinates of each central point and the length and angle of each row. In this way, the inter-row distance is varying for some pixels, and the field edges are not exactly straight. Finally, holes are randomly placed in order to simulate errors in the occupancy grid generation, and each image is randomly rescaled to get masks of different sizes. The actual row points are generated as filled circles with a random radius of 1 or 2 pixels, to address the possible variations the width of the rows.

\begin{figure}
\centering
\includegraphics[scale=0.4]{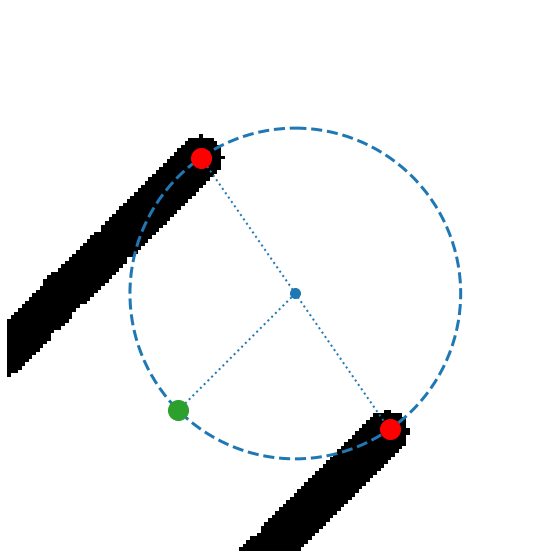}
\caption{To compute the waypoint location (green), we consider the circle of radius the semi-distance between the extremities of two adjacent rows and we find the point on the circumference in the direction given by the mean angle of the rows.}
\label{fig:wp_gen}
\end{figure}

\begin{figure*}
\centering
\begin{subfigure}{0.3\textwidth}
  \centering
  \includegraphics[width=0.85\linewidth]{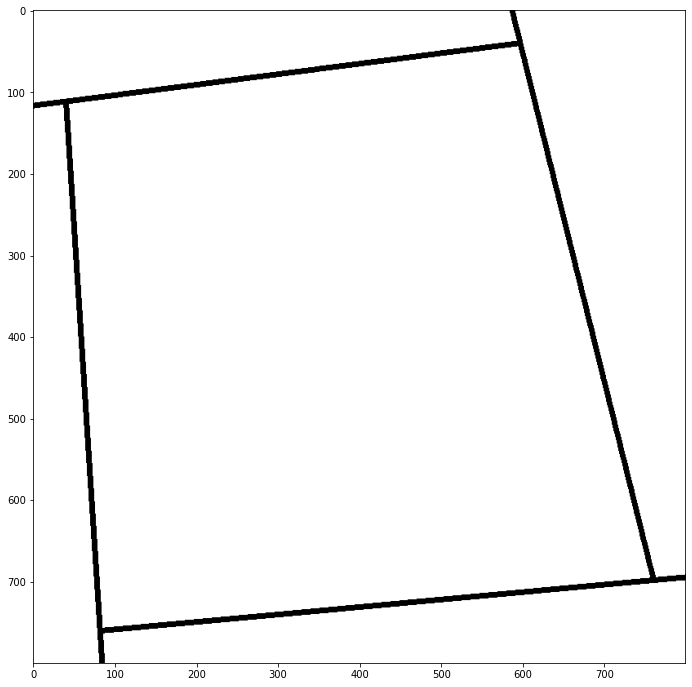}
\end{subfigure}
\begin{subfigure}{0.3\textwidth}
  \centering
  \includegraphics[width=0.85\linewidth]{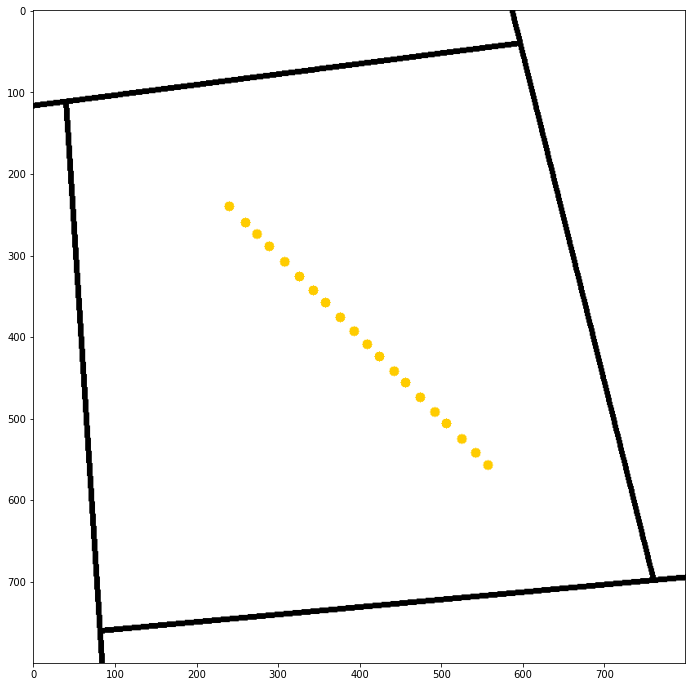}
\end{subfigure}
\begin{subfigure}{0.3\textwidth}
  \centering
  \includegraphics[width=0.85\linewidth]{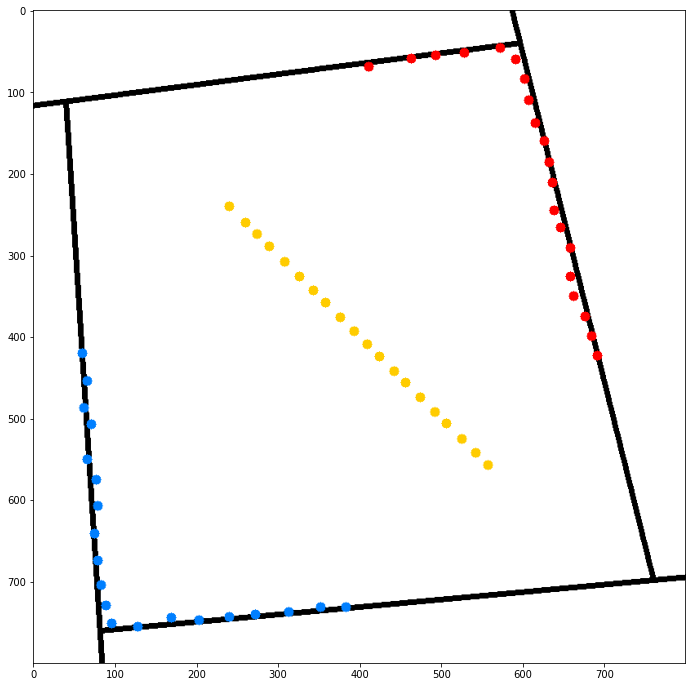}
\end{subfigure}
\begin{subfigure}{0.3\textwidth}
  \centering
  \includegraphics[width=0.85\linewidth]{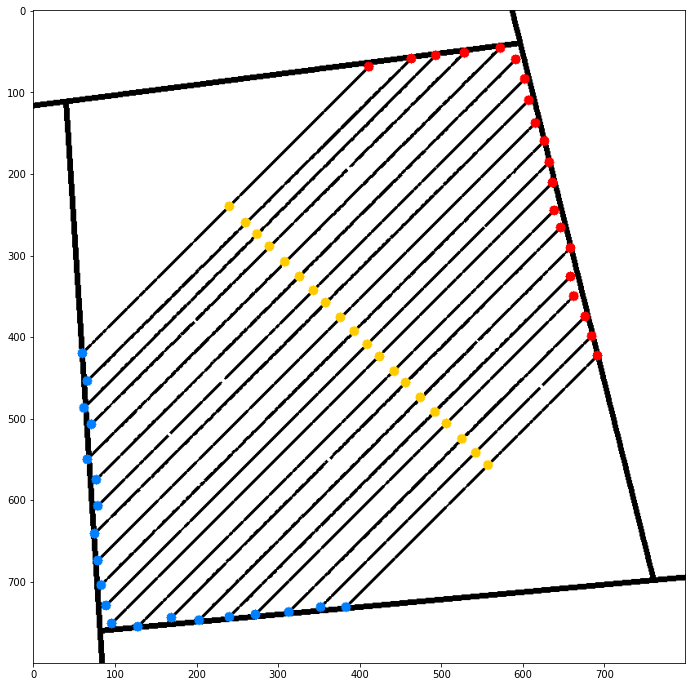}
\end{subfigure}
\begin{subfigure}{0.3\textwidth}
  \centering
  \includegraphics[width=0.85\linewidth]{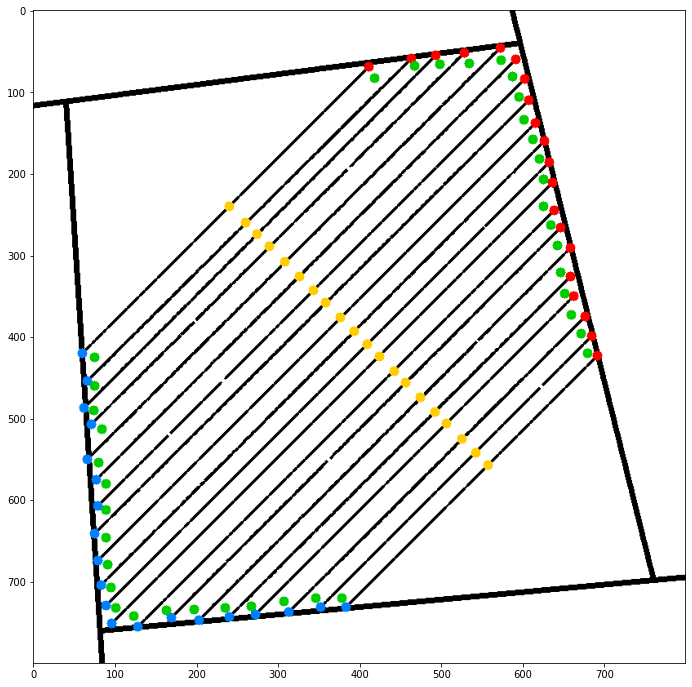}
\end{subfigure}
\begin{subfigure}{0.3\textwidth}
  \centering
  \includegraphics[width=0.85\linewidth]{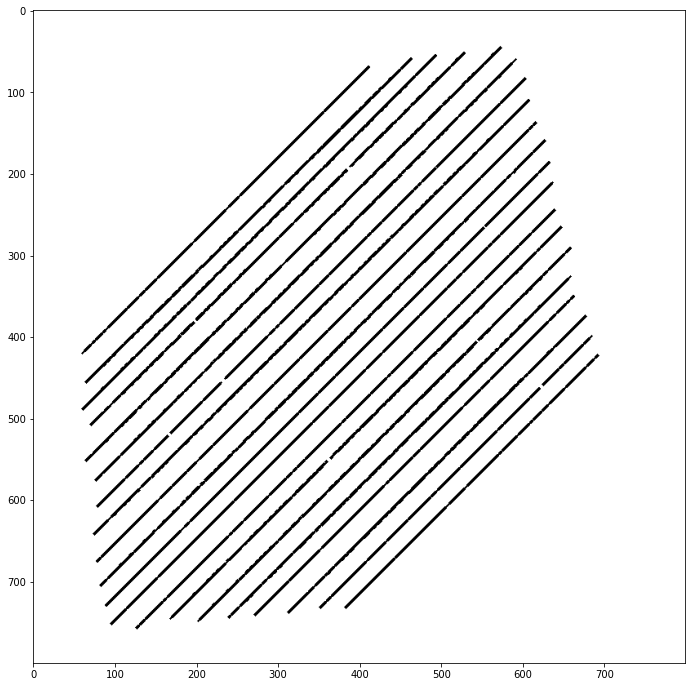}
\end{subfigure}
\caption{Occupancy grid generation process for a $800\times800$ mask with $N=20$ and $\alpha=\nicefrac{\pi}{4}$. Firstly random borders are generated, then, $N$ row centers (yellow) are identified starting from the image center. Starting (blue) and ending (red) points are found at the intersection with the borders, with some random displacement to add variability. The actual row lines are then generated, adding holes with a certain probability. Finally, the target waypoints (green) are found with the method presented in Fig. \ref{fig:wp_gen}.}
\label{fig:mask_generator}
\end{figure*}

To generate the ground truth waypoints, we start considering the mean between each pair of first and last points of the rows. Then, we move those points towards the inside of the field, , ensuring that waypoints are in between the two rows. That is a relevant aspect to ease the final path generation. Indeed, external waypoints could easily lead to wrong trajectories skipping some rows or going through some already covered. Fig. \ref{fig:wp_gen} shows how waypoints are generated and Fig. \ref{fig:mask_generator} illustrates all the steps for the masks generation.

\begin{figure*}
\centering
\begin{subfigure}{0.24\textwidth}
  \centering
  \includegraphics[width=\linewidth]{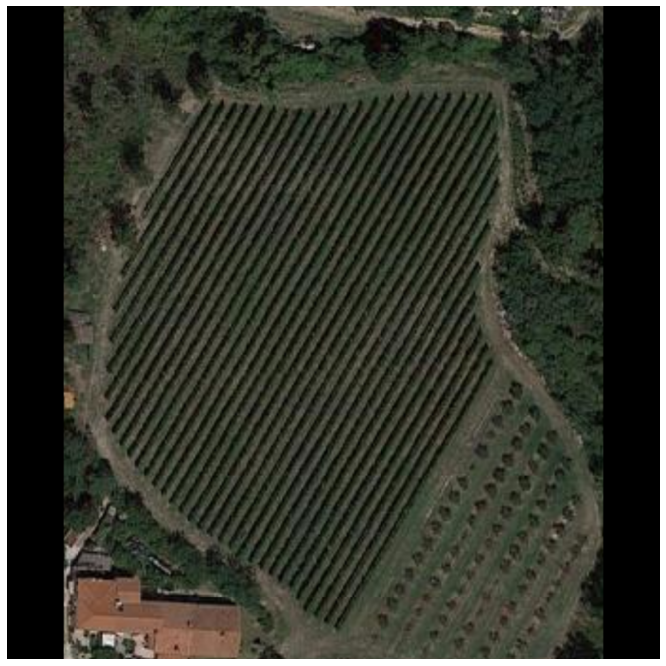}
\end{subfigure}
\begin{subfigure}{0.24\textwidth}
  \centering
  \includegraphics[width=\linewidth]{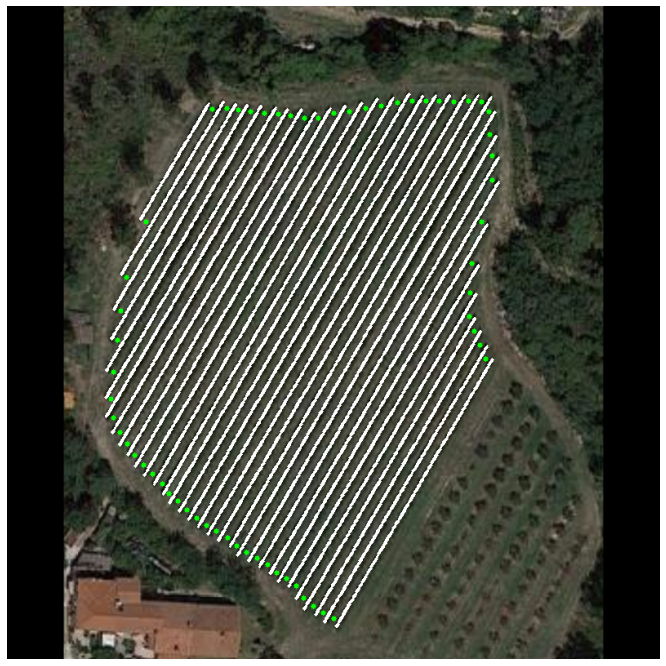}
\end{subfigure}
\begin{subfigure}{0.24\textwidth}
  \centering
  \includegraphics[width=\linewidth]{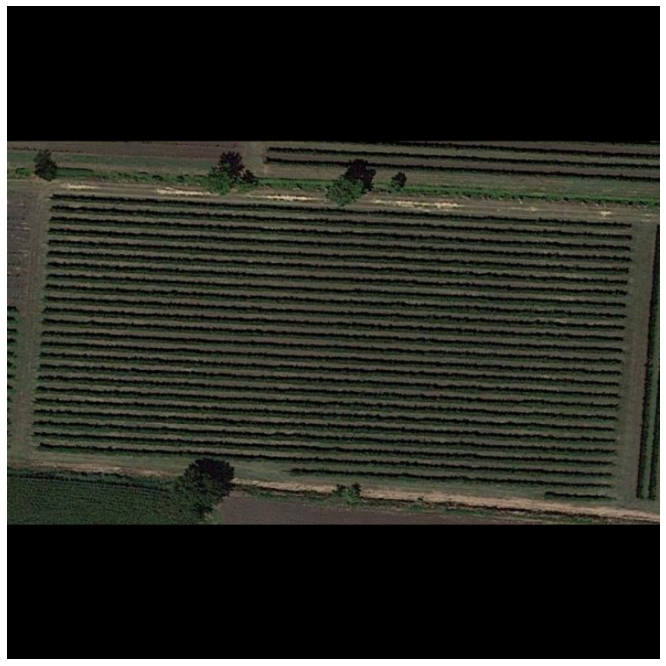}
\end{subfigure}
\begin{subfigure}{0.24\textwidth}
  \centering
  \includegraphics[width=\linewidth]{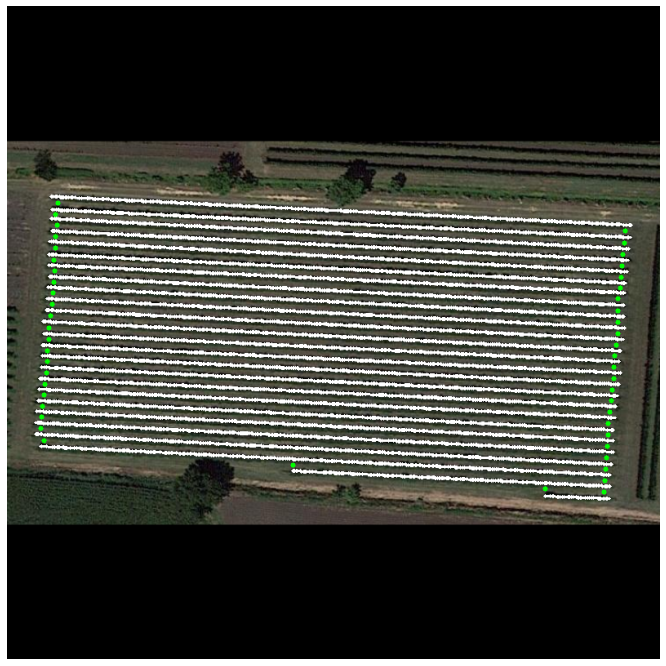}
\end{subfigure}
\caption{Two examples of real-world images taken from Google Maps satellite database and manually annotated. Green points are the ground truth waypoints computed with the method presented in Fig. \ref{fig:wp_gen}.}
\label{fig:satellite}
\end{figure*}

In addition to the synthetically generated dataset, we also manually collect and annotate 100 satellite images of different row-based crop scenarios from the Google Maps database. Those images are manually processed to extract both the occupancy grid and the target waypoints for the prediction evaluation. On average, the collected images have a ratio of 0.243 meters for pixel, with an inter-row average distance of 2.61 m. Real-world images are essential to demonstrate the ability of our approach to generalize to real-case scenarios and that training the network with a synthetic dataset is equally effective. Fig. \ref{fig:satellite} shows two examples of manually annotated satellite images used as the test set. It is worth to mention that the satellite images also present different variations, with respect to the ones present in the synthetic dataset (e.g. slight row curvature), in order to test the robustness of the methodology.

\section{Methodology}
Given an occupancy grid map of the row crop taken into account, we frame the end rows waypoint detection as a regression problem, estimating positions of the different points with a fully convolutional deep neural network. So, a single neural network, DeepWay, predicts the global path waypoints directly from the full input image, straight from image pixels to points in one evaluation. Since the whole detection pipeline is a single model, it can be optimized end-to-end directly on waypoints estimation. Prior works, on global path generation for row-based crops, heavily rely on local geometric rules and hardcoded processes that struggle to scale and generalize to the variability of possible real scenarios. On the other hand, DeepWay learns to predict all waypoints of the input crop's grid map simultaneously together with their corrections using features coming from the entire image. It trains on full occupancy grid maps optimizing directly waypoints estimation performance and reasoning globally about the input data. Finally, a post-processing waypoint refinement and ordering algorithm is used to correct missing points, misplaced detections and order them before the final global path generation.  
\subsection{Waypoint estimation}
\label{waypoint_estimation}
\begin{figure*}
\centering
\includegraphics[scale=0.50]{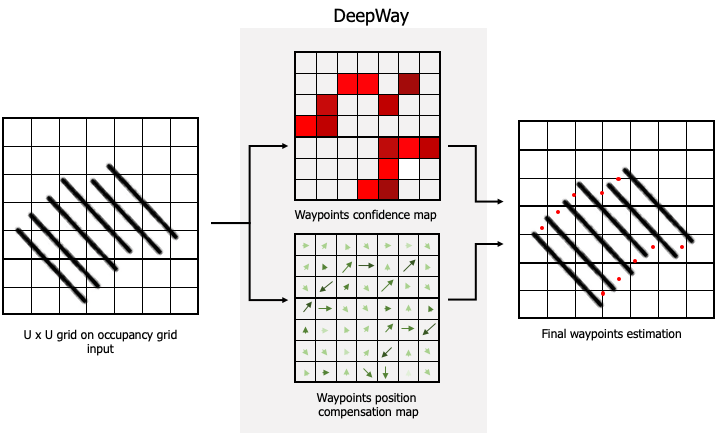}
\caption{DeepWay models waypoints detection as a regression problem. It analyzes the input occupancy grid map with a grid of $U_{h} \times U_{w}$ and for each grid cell it predicts a waypoint confidence probability $P(\text{wp})$ and the correspondence coordinate compensations $\Delta_{x}$, $\Delta_{y}$. In the scheme, the grid has equal dimensions $U$ for both axes.}
\label{fig:deepway_ex}
\end{figure*}

Our methodology divides the input image of dimension $H \times W$ into a $U_{h} \times U_{w}$ grid and if the centre of an end row waypoint falls into a grid cell, that cell is responsible for detecting that point. Each cell, $u$, predicts a confidence probability $P(u)$ that reflects how confident is the network that a waypoint is placed inside its boundaries. If no waypoint is present, the confidence score should tend to zero. Otherwise, we want to be more close to one as possible.
Moreover, each grid cell predicts a position compensation couple $(\Delta_{x} , \Delta_{y})$ that, if necessary, moves the predicted points from the centre of the cell. In Fig. \ref{fig:deepway_ex} is presented a high-level overview of the operation principle of the methodology. DeepWay, given an occupancy grid input, produces a confidence map of dimension $U_{h} \times U_{w}$ with probability $P(u)$ for each cell $u$ with their relative corrections $(\Delta_{x} , \Delta_{y})$. Either output maps are used by a simple interpret function to produce the final waypoints estimation on the original occupancy grid input.

\begin{figure}
\includegraphics[scale=0.16]{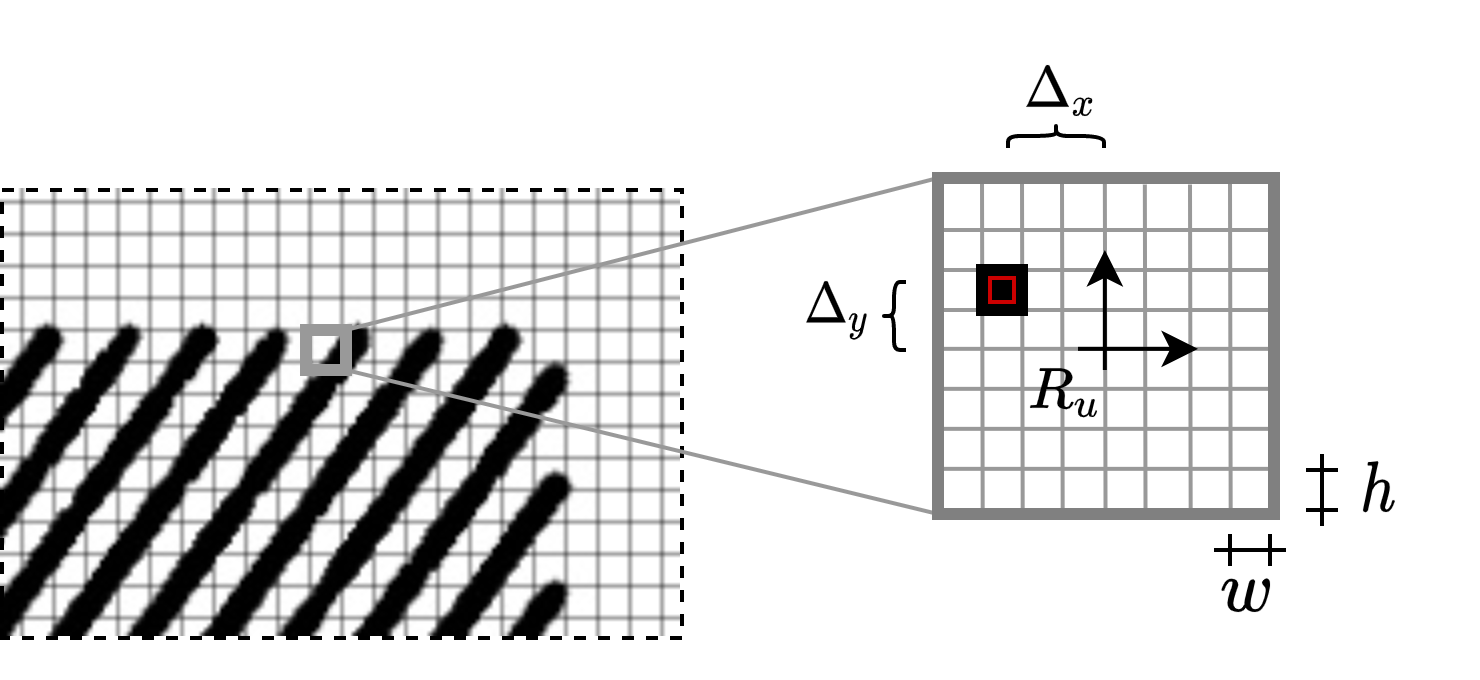}
\caption{DeepWay estimates for each cell $u$ a probability $P(u)$ and a position compensation couple $(\Delta_{x} , \Delta_{y})$ to better adjust detected waypoints on the original occupancy map dimension, $H \times W$. The highlighted area shows with a red square the actual position of the specific ground truth and the need to displace the prediction from the centre of the cell.}
\label{fig:grid_ex}
\end{figure}

Indeed, the inference grid $U_{h} \times U_{w}$ is $k$ times smaller than the original dimensions, $H$ and $W$, of the input. So, each $u$ cell contains $k \times k$ original pixels. Without an explicit position compensation mechanism, the network would not be able to adjust the position of a waypoint detection, being unable to place it in the correct position of the original input space dimension. As depicted in Fig. \ref{fig:grid_ex}, where the $U_{h} \times U_{w}$ is superimposed to the occupancy grid input, most of the row terminal parts do not have a centred $u$ cell that can perfectly fit a prediction. Indeed, as in the case of the highlighted area, two $u$ cells cover the specific end row, and none of the two perfectly fits the position of the ground-truth placed in the middle point that connects the two side rows. Nevertheless, each cell contains $k \times k$ positions that can be used to refine the placement of an eventual waypoint detection. More specifically, each $u$ grid cell can predict two values, $\Delta_{x}$ and $\Delta_{y}$, that let displace possible prediction respect to a reference $R_{u}$ placed in in the centre of the cell. So, the coordinates of a certainly detected waypoint in the original input dimension $H \times W$ can be found using the following equation:

\begin{equation}
\label{upres}
    \hat{\bold{y}}_{O}=k(\hat{\bold{y}}_{U} + \frac{\bold{\Delta + 1}}{2})
\end{equation}

where $\hat{\bold{y}}_{O}$ and $\hat{\bold{y}}_{U}$ are the two vectors containing the coordinates $x$ and $y$ in the $\mathcal{R}_{O}$ and $\mathcal{R}_{U}$ reference frames, respectively. Position compensations are normalized, and the reference frame, $\mathcal{R}_{u}$, of the cell $u$ is centred respect to the cell itself.

Therefore, in order to obtain the final waypoints estimation in the original input space, a confidence threshold $t_{c}$ is applied to the waypoints confidence map in order to select all detected waypoints with a probability $P(u)>t_{c}$. Furthermore, Eq. \ref{upres} is used on all selected waypoints in conjunction with the position compensation maps in order to obtain the respective coordinates on the original reference frame of the input. Finally, a waypoint suppression algorithm is applied to remove all couple points with a reciprocal Euclidian distance inferior to a certain threshold $d_{c}$. The predicted waypoint with the highest $P(u)$ is maintained and the remaining ones discarded.

\subsection{Network Design}
\begin{figure*}
\centering
\includegraphics[scale=0.50]{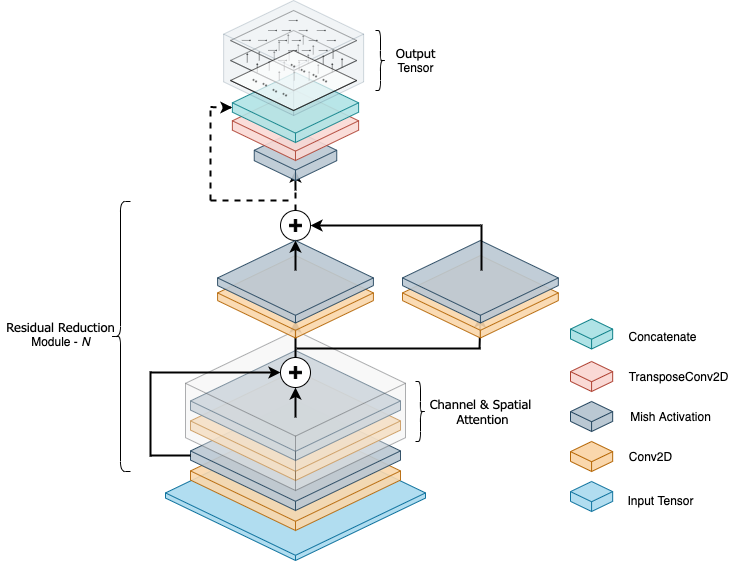}
\caption{Overview of the DeepWay architecture. The model takes a tensor $\mathrm{\bold{X}}^{(i)}$ as input and reduces its spatial dimension with a stack of $N$ residual reduction modules. The synergy of the channel and spatial attention layers let the network focus on more promising and relevant features. Finally, the neural network outputs a tensor $\mathrm{\bold{Y}}^{(i)}$ of dimension $U \times U \times 3$ with probability $P(u)$ and position compensation couple $(\Delta_{x} , \Delta_{y})$ for each cell $u$.}
\label{fig:deepway_architecture}
\end{figure*}
DeepWay is a fully convolutional neural network that is directly fed with an occupancy grid map of a row-based crop and predicts waypoints for a successive global path generation. In particular, an input tensor $\mathrm{\bold{X}}^{(i)}$ is progressively convoluted by a stack of $N$ residual reduction modules. Each module is composed of a series of convolutional 2D layers with Mish, \cite{misra2019mish}, as activation function and channel and spatial attention layers to let the network highlight more relevant features, \cite{woo2018cbam}. Moreover, each module terminates with a convolutional layer with stride two in order to reduce the spatial dimension of the input tensor. After $N$ residual reduction modules, the two first dimensions are reduced by a factor $k+1$. Therefore, a transpose convolutional layer with stride two is interposed in order to obtain an output tensor with the two first dimension equal to $U \times U$. Moreover, as firstly introduced by segmentation networks, \cite{ronneberger2015u},  a residual connection with the output tensor coming from the $N-1$ block is added in order to include important spatial information to the tensor before the last layer. Finally, similarly to single-stage object detection network \cite{redmon2016you,liu2016ssd}, the output tensor $\mathrm{\bold{\hat{Y}}}^{(i)}$ with shape $U_{h} \times U_{w} \times 3$ is computed with a 1x1 convolution operation with sigmoid and tanh as activation functions for the first and the two last channels, respectively. Indeed, sigmoid allows obtaining a confidence probability $P(u)$ predicting the presence of a possible waypoint. On the other hand, the tanh function, being limited between -1 and +1, computes the two coordinate compensations $\Delta_{x}$, $\Delta_{y}$ for each cell. Finally, the post-processing pipeline discussed in Section \ref{waypoint_estimation} is used to process the output tensor further $\mathrm{\bold{\hat{Y}}}^{(i)}$ and obtain the final waypoints estimation in the original input space.
 An overview of the overall architecture of the network is shown in Fig. \ref{fig:deepway_architecture}.

\subsection{Waypoint Refinement and Path Generation}
In order to generate a suitable path from the waypoints, we further process the network predictions to refine them and identify the correct order for connecting the waypoints. We cluster the predicted points using the density-based clustering algorithm DBSCAN \cite{ester1996density}. This approach allows to automatically cluster together points that are close to each other and can give a first subdivision of the waypoints into main groups. Depending on the geometry of the field, several clusters can be found in this way, and some points can remain unclustered, in particular for rows drastically shorter with respect to the others. To get the order of the waypoints inside each group, we project each of them along the perpendicular to the rows, and we sort them in this new reference system.

The row angle is estimated with the progressive probabilistic Hough transform technique \cite{matas2000robust}. This algorithm is a classic computer vision feature extraction method, able to detect lines in an image and return an estimate of starting and ending points. Even though this algorithm may seem enough to solve the whole problem of finding the target waypoints in the mask without the need of a neural network, this approach is too dependent on a number of hyper-parameters that cannot be well-defined a-priori and generally is not able to cope with holes and irregularities which are inevitably present in real-world field occupancy grids. We experimentally find that the application of this method leads to a high number of false-positive and false-negative detections of lines on both the synthetic and the satellite datasets. However, we still use it to estimate the row angle by averaging the orientations of each couple of detected points. In the case of a complete failure of this approach that can happen with the most complex masks, we estimate the angle using a probabilistic iterative process that minimizes the number of intersections with the rows starting from points close to the image centre. 

After ordering the points inside each cluster, we adopt a refinement approach to insert possible missing waypoints or deleting duplicated ones, by counting the number of rising and falling edges in the mask along the line connecting two consecutive points. Then, to get the final order, the different clusters must be merged into two groups A and B containing the waypoints at the beginning and at the end of each row. We adopt a strategy to iteratively assign clusters to the groups considering their size and the values of their projections along the perpendicular to the rows. We assume that a good assignment is the one that spans the same interval along the projection axis on both the groups with different clusters. After the assignments, we refine the borders between the merged clusters, in order to compensate for possible mispredicted points. Once we get the final groups, we compute the order by considering a pattern A-B-B-A. Every intra-groups connection is performed by checking possible intersections with the rows and correcting the order consequently. If there is a missing point in one of the two groups even after the waypoints refinement process, we remain within the same group, avoiding any possible intersection with the rows. In this way, we put the focus on building feasible paths in the field.

To compute the global path, we use the A* search algorithm \cite{hart1968formal} from waypoint to waypoint following the pre-computed order. In particular, at each iteration, the algorithm chooses the path that minimizes the following cost function:  
\begin{equation}
\label{astar}
    f(n) = g(n)+  w \cdot h(n)
\end{equation}
where \emph{n} is the next step on the path, \emph{g(n)} is the cost of the path from the previous waypoint to \emph{n}, and \emph{w} is the weight given to the heuristic function \emph{h(n)} that estimates the cost of the cheapest path from \emph{n} to the next waypoint. In particular, as heuristic function, the euclidean distance is proved to be more time-efficient than the Manhattan distance \cite{Craw2017}, and, since the trajectory is quite straight, we assign a relevant weight \emph{w} to speed up the path generation to achieve better performance.
The A* search algorithm is proved to be the simplest and generic ready-to-use path generation algorithm to produce the global output trajectory \cite{costa2019survey}. Nevertheless, our experimentation pointed out that further efforts should be dedicated to developing a more tailored solution that exploits the prior knowledge of the row-based environment and post-processing derived information in order to boost the run-time efficiency and to output a more centred path along the rows.

\begin{figure*}
\centering
\begin{subfigure}{0.48\textwidth}
  \centering
  \fbox{\includegraphics[width=0.85\linewidth]{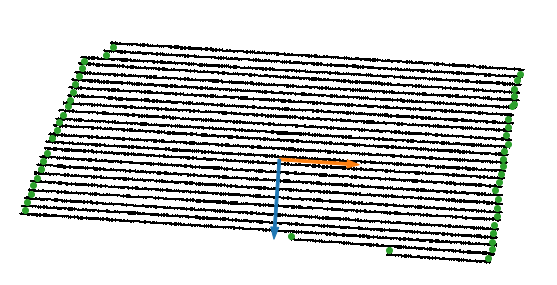}}
\end{subfigure}
\begin{subfigure}{0.48\textwidth}
  \centering
  \fbox{\includegraphics[width=0.85\linewidth]{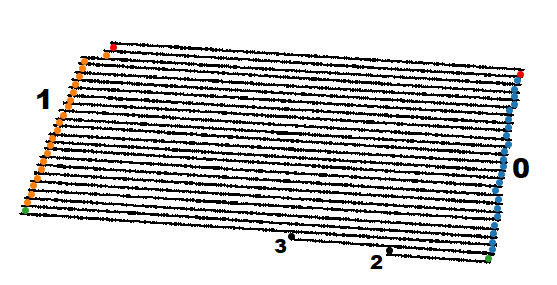}}
\end{subfigure}
\vskip 7pt
\begin{subfigure}{0.48\textwidth}
  \centering
  \fbox{\includegraphics[width=0.85\linewidth]{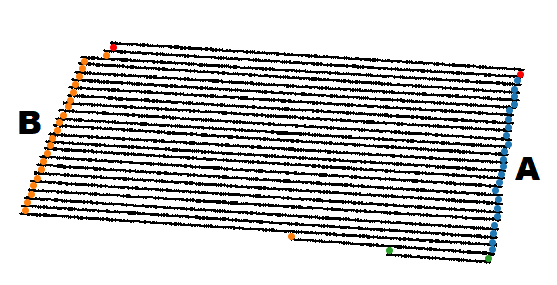}}
\end{subfigure}
\begin{subfigure}{0.48\textwidth}
  \centering
  \fbox{\includegraphics[width=0.85\linewidth]{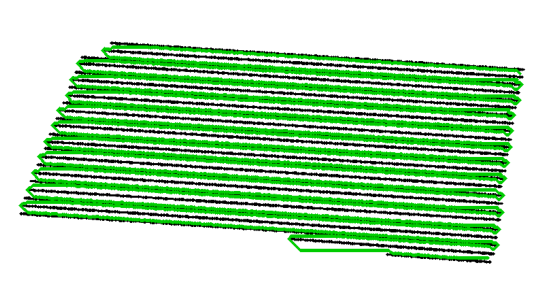}}
\end{subfigure}
\caption{Waypoints refinement and path generation process. Firstly, the row angle is estimated. Then, the predicted waypoints are clustered with the DBSCAN algorithm \cite{ester1996density}. We iteratively merge the clusters into two principal groups A and B and finally we obtain the final order of the predicted waypoints. Applying a global path generation method like the A* search algorithm \cite{hart1968formal}, it is possible to get the final path, represented in green.}
\label{fig:postprocessing}
\end{figure*}

Fig. \ref{fig:postprocessing} shows all the operations performed during the waypoints refinement and ordering process and the global path computation. The full pipeline of the proposed approach is presented in Alg. \ref{alg:full_algorithm}.

\begin{algorithm}[b]
\SetKwInOut{Input}{input}\SetKwInOut{Output}{output}
\SetKwData{Wp}{wp}\SetKwData{Wpc}{wp$|_c$}
\SetKwData{Clust}{clusters}\SetKwData{Proj}{proj}
\SetKwData{Ord}{order}\SetKwData{A}{A}\SetKwData{B}{B}
\SetAlgoLined
 \Input{Occupancy grid $X$ of size $H\times W$}
 \Output{Path $p$}
 \BlankLine
 \Wp$\gets$ DeepWay($X$)\;
 $\alpha\gets$ angle\_estimation($X$)\;
 \Clust $\gets$ DBSCAN(\Wp)\;
 
 \ForEach{cluster c}{
    \Proj $\gets$ project(\Wpc,$\alpha$)\;
    \Ord $\gets$ sort(\Proj)\;
    \Wpc$\gets$ refine(\Wpc,\Ord,$X$)\;
 }
 \A,\B $\gets []$\;
 \While{\Wp is not empty}{
    \A,\B $\gets$ assign(\Wp,$\alpha$,\A,\B)\; 
 }
 $p\gets$ final\_order(\A,\B,$X$)\;
 $p\gets$ A*($p,X$)\;
\caption{Global path generation pipeline}
\label{alg:full_algorithm}
\end{algorithm}

\section{Experiments and Results}
\label{sec:results}

\subsection{Training}
The first convolutional layer and the last one have 7 and 3 as kernel sizes. All other ones have 5 and 16 as kernel dimension and number of filters, respectively. On the other hand, we adopt the same default parameters for the channel and spatial attention layers of Woo et al. \cite{woo2018cbam}. We use 3000 synthetic images for training with a resolution of 800x800 and $k=8$. So, each prediction made by the network, $\mathrm{\bold{\hat{Y}}}^{(i)}$, over a grid $U_{h} \times U_{w}$ with equal axis size, before the post-processing, has a spatial dimension of 100x100. Moreover, we train the network with 200 epochs using Adam optimizer, \cite{kingma2014adam}, with a fixed learning rate equal to $\eta=3e-4$ and batch size of 16. The optimal learning rate, $\eta$, is experimentally derived using the methodology described in \cite{smith2017cyclical}.
\begin{equation}
\label{loss_function}
    J(\Theta)=\sum_{i,j=0}^{U}[\mathbbm{1}^{wp}_{i,j}\lambda^{wp}(y_{i,j}-\hat{y}_{i,j})^{2}+
    \mathbbm{1}^{nowp}_{i,j}\lambda^{nowp}(y_{i,j}-\hat{y}_{i,j})^{2}]
\end{equation}
Finally, we use the loss function of Eq. \ref{loss_function} that is a modified version of the $L_{2}$ sum-squared error. $\mathbbm{1}^{wp}_{i,j}$ and $\mathbbm{1}^{nowp}_{i,j}$ denote if a waypoint is present or absent from the $i,j$ cell. Therefore, it is possible to give more relevance to cells with a waypoint that are considerably less than true negatives. After a grid search analysis, $\lambda^{wp}$ and $\lambda^{nowp}$ are set to 0.7 and 0.3, respectively. That solution stabilizes training preventing to overpower the gradient from cells
that do contain a waypoint.

The resulting fully-convolutional network is a light-weight model with less than 60,000 parameters, a negligible inference latency and that can be easily trained with a commercial GPU in less than 20 minutes. We make use of a workstation with an NVIDIA 2080, 8 GB of RAM and the TensorFlow 2.x machine learning platform \cite{abadi2016tensorflow}.

\subsection{Waypoint Estimation Evaluation}
\begin{figure*}
\centering
\begin{subfigure}{.32\textwidth}
  \centering
  \includegraphics[scale=0.37]{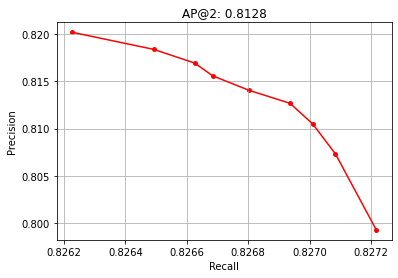}
  \caption{}
  \label{fig:ap_fig1}
\end{subfigure}%
\begin{subfigure}{.32\textwidth}
  \centering
  \includegraphics[scale=0.37]{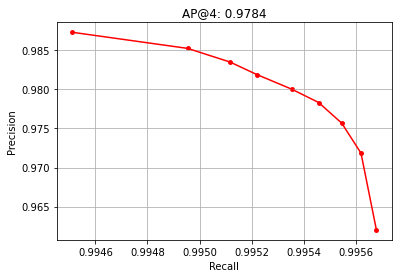}
  \caption{}
  \label{fig:ap_fig2}
\end{subfigure}
\begin{subfigure}{.32\textwidth}
  \centering
  \includegraphics[scale=0.37]{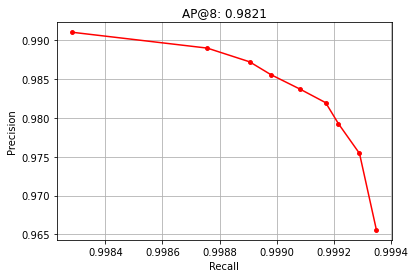}
  \caption{}
  \label{fig:ap_fig3}
\end{subfigure}
\caption{Average Precision results on the synthetic test set for different values of $r_{c}$. More restrictive ranges obtain lower values of recall and precision. Conversely, decreasing the value of $t_{c}$ the precision is mostly affected due to the growing number waypoints with low-score that are predicted by the network.}
\label{fig:ap_fig}
\end{figure*}
After training, the network is evaluated with 1000 synthetic images. The evaluation aims at assessing its precision and recall in detecting points within a certain radius $r_{c}$ from the ground-truth. Moreover, as explained in \ref{waypoint_estimation}, the waypoint estimation is found setting a certain value of a confidence threshold $t_{c}$. So, different values of recall and precision can be obtained fixing different thresholds. For that reason, we adopt an adaptation of the Average Precision (AP) metric that is commonly used across object detection challenges like PASCAL Visual Object Classes (VOC), \cite{everingham2010pascal}, and MS Common Objects in Context (COCO), \cite{lin2014microsoft}. Therefore, if a waypoint prediction is within the selected radius $r_{c}$ is counted as a true positive (TP). However, if more predictions fall within the selected range, only one is counted as TP and all others as false positive (FP). On the other hand, all ground-truth not covered by a prediction are counted as false negatives (FN). So, the AP computation at a certain distance $r_{c}$ is obtained with common definition of recall and precision, varying the value of threshold $t_{c}$ from zero to one. We set the distance threshold for the waypoints suppression to $d_c=8$ pixels. 
In Fig. \ref{fig:ap_fig} are depicted three graphs obtained with different values of $r_{c}$ and 0.1 as size step for $t_{c}$. At a distance range of 8 pixels the Average Precision is equal to 0.9821, but with only a radius of 2 the AP value drops at 0.8128. Moreover, as expected by this metric, recall and precision are inversely proportional by modifying the value of the threshold $t_{c}$.
\begin{figure}
\centering
\includegraphics[scale=0.27]{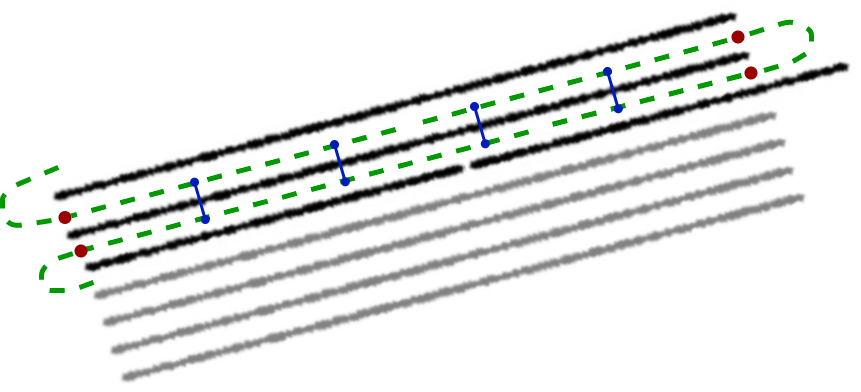}
\caption{Scheme of the working principle of the algorithm to calculate the coverage score metric. The equally spaced points are represented in blue, the planned path in green, and in black the crop rows from the occupancy grid. The blue lines are the segments used to check eventual intersections in the occupancy grid.}
\label{fig:CoverAlgo}
\end{figure}

In addition to the synthetic test dataset, we also compute the AP metric on the manually annotated satellite images. We reach an AP of 0.9794 with a distance range $r_c$ of 8 pixels (1.94 m on average), 0.9558 with 4 (0.97 m on average) and 0.7500 with 2 (0.49 m on average). As expected, these results are slightly worse with respect to the synthetic images when with low values of $r_c$, since the real-world masks are generally more complex, with irregular borders and with sudden changes in  the length of the rows. All these aspects are only marginally covered by our synthetic generation process, but this does not cause a high drop in the AP metric, meaning that our approach is able to generalize to real-world examples with high-quality predictions. 

\subsection{Path Generation Evaluation}
\begin{figure*}
\centering
\begin{subfigure}{0.45\textwidth}
  \centering
  \includegraphics[width=0.7\linewidth]{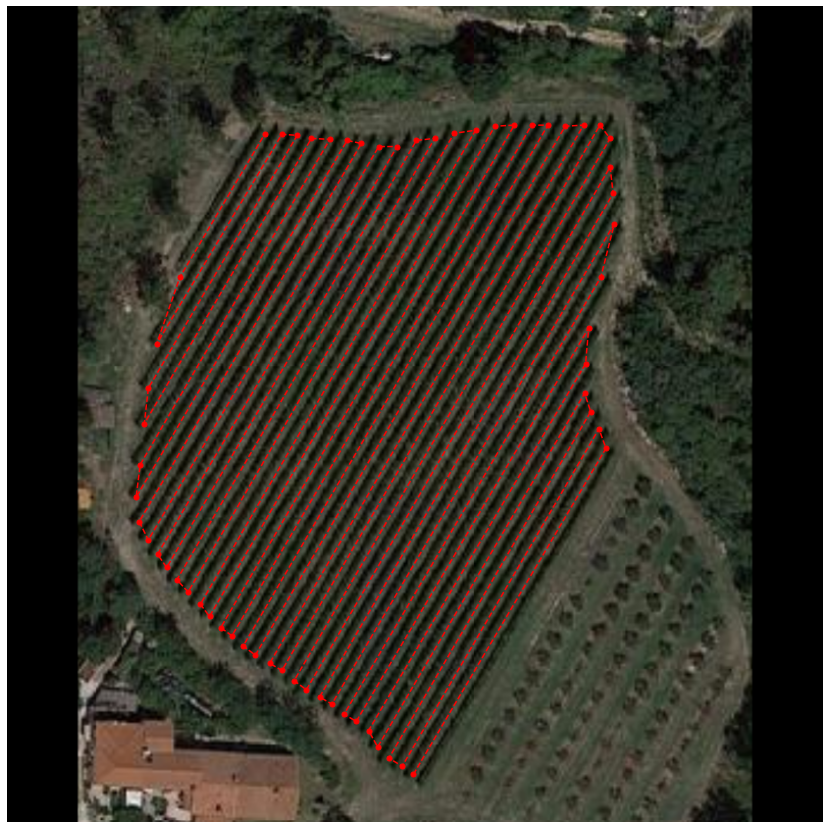}
\end{subfigure}
\begin{subfigure}{0.45\textwidth}
  \centering
  \includegraphics[width=0.7\linewidth]{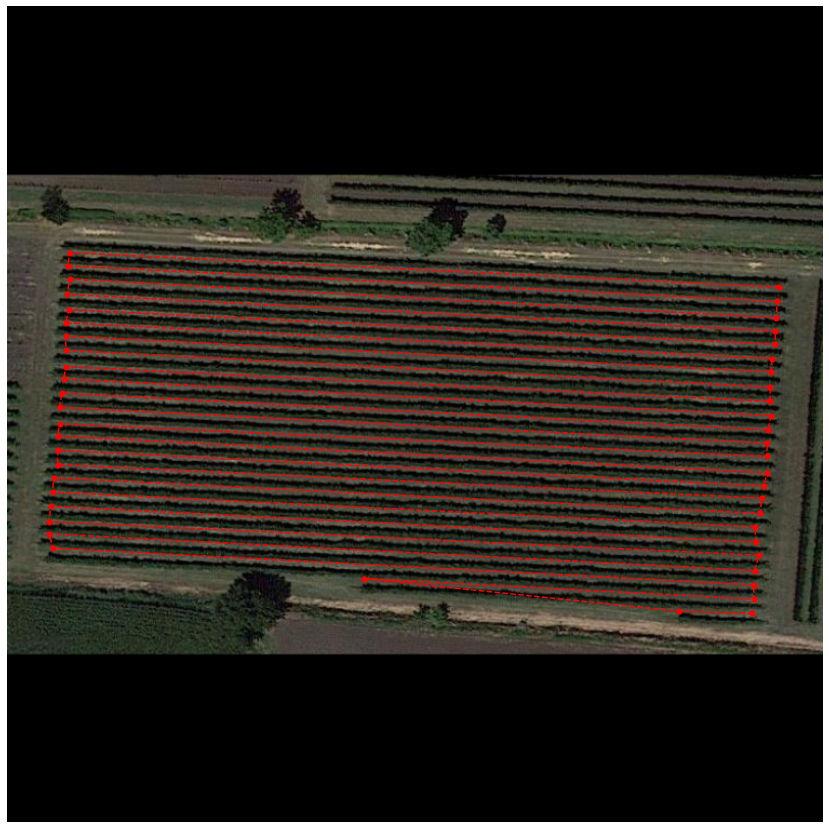}
\end{subfigure}
\begin{subfigure}{0.45\textwidth}
  \centering
  \includegraphics[width=0.7\linewidth]{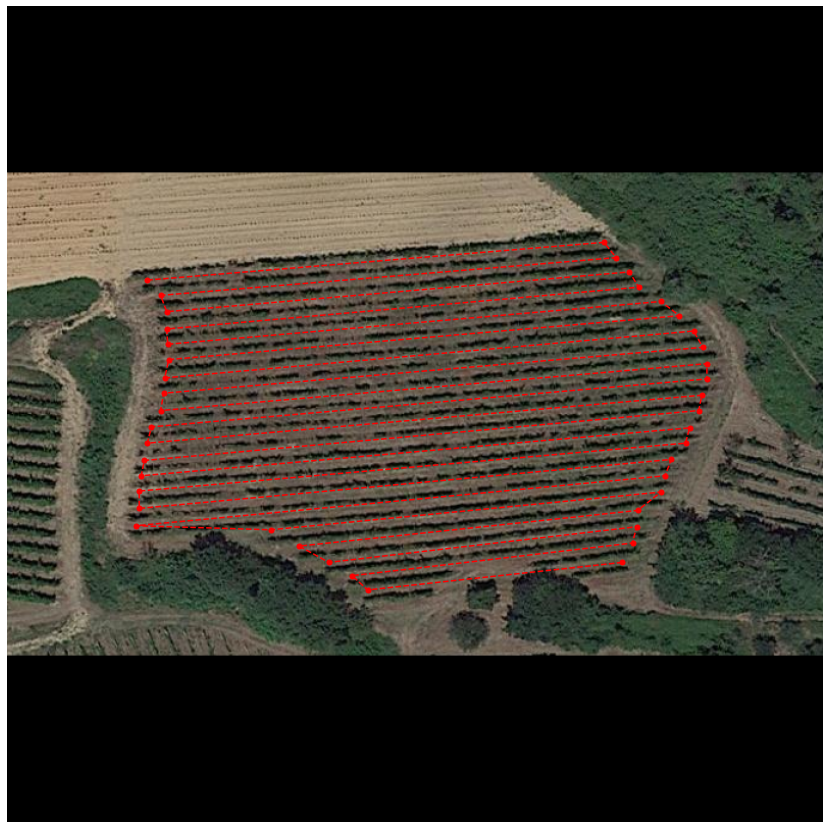}
\end{subfigure}
\begin{subfigure}{0.45\textwidth}
  \centering
  \includegraphics[width=0.7\linewidth]{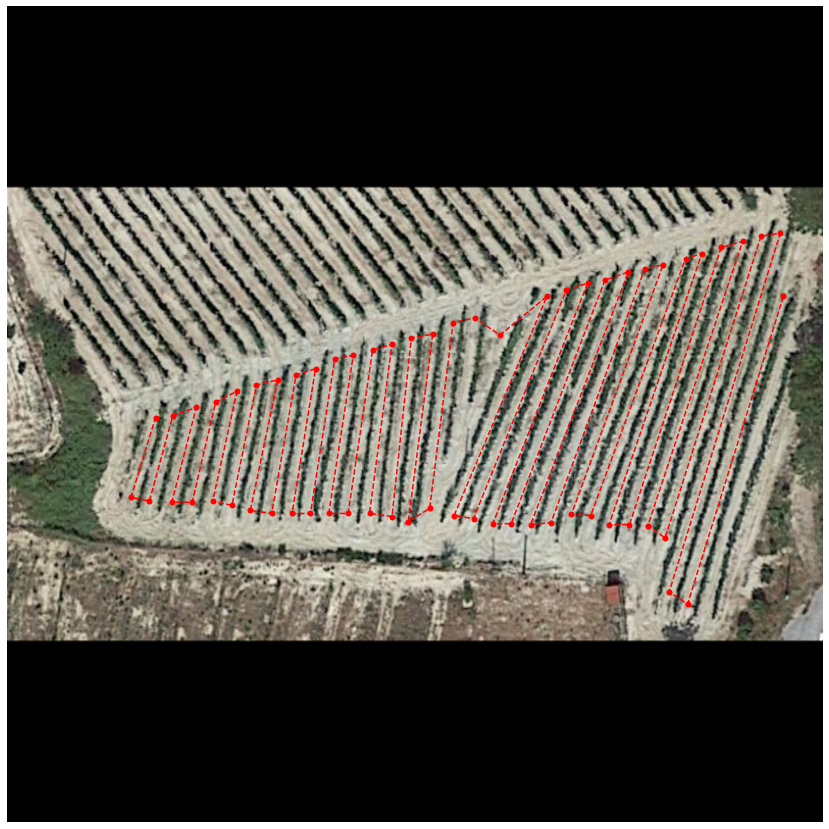}
\end{subfigure}
\caption{Some examples from the dataset of real-world satellite images taken from Google Maps with the ordered predicted waypoints.}
\label{fig:results}
\end{figure*}
In order to assess a coverage percentage of a generated global path, we define a coverage score (CS) metric to be computed for each instance crop $i$ as follow:
\begin{equation}
\label{CoveragePrecisionFunct}
    CS^{(i)}= \frac{\mbox{number of correctly covered crop rows}}{\mbox{total number of crop rows}}
\end{equation}
where a row is considered correctly covered if the generated trajectory passes just one time along the entire row.

The algorithm to compute the CS metric is briefly schematized in Fig. \ref{fig:CoverAlgo}. It chooses four equally spaced points in the A* planned path between two consecutive waypoints that do not belong to the same cluster. Successively, each point is connected to the corresponding point of the next row in order to create a segment. Checking the occupancy grid of the parcel, if at least one of the segment intersects one crop row, it means that the row is correctly covered, otherwise, in case of zero or more than one intersected rows, the row is not correctly covered. The algorithm is iterated for the whole parcel, and the coverage score is computed.

We reach a mean coverage of 0.9648 on the synthetic test set and 0.9409 on the real-world test set. Both values are obtained with a score threshold of $t_c=0.9$, selected using AP curves, in order to increase the predictions precision and a distance threshold $d_c=8$ pixels for the waypoints suppression. Since the refinement process is able to add missing points in clusters, it is better to pay in recall and ensure high precision, for the final path generation process. In general, we find that incomplete coverages are mainly caused by a too-small inter-row distance that causes less quality prediction and by too short rows that can cause lousy clustering and therefore final row skipping. DeepWay is a fully-convolutional network and, so better results could be achieved, increasing the resolution of the input tensor at inference time. That could lead to more considerable inter-row distance and higher precision. Further works will investigate also mixed resolutions training in order to enhance this capability of the network. Nevertheless, the obtained results, underline how our approach can be successfully used to plan a global path for automatic navigation inside a row-crop field. 

\subsection{Qualitative results}
In Fig. \ref{fig:results}, as output examples of the proposed approach, we present a set of satellite images with the corresponding predicted and ordered waypoints. In the third image, it is possible to observe how our approach is also able to handle big holes in the rows. On the other hand, the fourth image shows how fields with variable orientations can cause sub-optimal predictions and incomplete field coverage.

An important aspect to be underlined, is that the proposed methodology allows to perform an automatic global path generation with precise location information, using geo-referenced satellite or drone images as input to the network. Therefore, it can be used for real-time navigation and localization of autonomous robots inside the fields without the need for manually identify and measure the GPS coordinates of navigation points.

\section{Conclusions}
We introduce DeepWay, a novel representation learning solution to tackle the global automatic path planning generation for row-based crops. The suggested algorithm is a light-weight and robust solution wrapped around a deep learning model that automatically generates waypoints for a given occupancy grid map. Extensive experiments with our synthetic dataset and real-world remote sensing derived maps demonstrated the effectiveness and scalability of our proposed methodology.
Further works will aim at integrating DeepWay with a segmentation network in order to jointly compute waypoints with the related occupancy grid map from the remote sensing colour space.

\section*{Acknowledgements}
This work has been developed with the contribution of the Politecnico di Torino Interdepartmental Centre for Service Robotics PIC4SeR\footnote{https://pic4ser.polito.it} and SmartData@Polito\footnote{https://smartdata.polito.it}.


\bibliographystyle{IEEEtran}
\bibliography{mainArXiv}

\begin{thebibliography}{10}
\providecommand{\url}[1]{#1}
\csname url@samestyle\endcsname
\providecommand{\newblock}{\relax}
\providecommand{\bibinfo}[2]{#2}
\providecommand{\BIBentrySTDinterwordspacing}{\spaceskip=0pt\relax}
\providecommand{\BIBentryALTinterwordstretchfactor}{4}
\providecommand{\BIBentryALTinterwordspacing}{\spaceskip=\fontdimen2\font plus
\BIBentryALTinterwordstretchfactor\fontdimen3\font minus
  \fontdimen4\font\relax}
\providecommand{\BIBforeignlanguage}[2]{{%
\expandafter\ifx\csname l@#1\endcsname\relax
\typeout{** WARNING: IEEEtran.bst: No hyphenation pattern has been}%
\typeout{** loaded for the language `#1'. Using the pattern for}%
\typeout{** the default language instead.}%
\else
\language=\csname l@#1\endcsname
\fi
#2}}
\providecommand{\BIBdecl}{\relax}
\BIBdecl

\bibitem{kamilaris2017review}
A.~Kamilaris, A.~Kartakoullis, and F.~X. Prenafeta-Bold{\'u}, ``A review on the
  practice of big data analysis in agriculture,'' \emph{Computers and
  Electronics in Agriculture}, vol. 143, pp. 23--37, 2017.

\bibitem{kamilaris2018deep}
A.~Kamilaris and F.~X. Prenafeta-Bold{\'u}, ``Deep learning in agriculture: A
  survey,'' \emph{Computers and electronics in agriculture}, vol. 147, pp.
  70--90, 2018.

\bibitem{jhuria2013image}
M.~Jhuria, A.~Kumar, and R.~Borse, ``Image processing for smart farming:
  Detection of disease and fruit grading,'' in \emph{2013 IEEE Second
  International Conference on Image Information Processing (ICIIP-2013)}.\hskip
  1em plus 0.5em minus 0.4em\relax IEEE, 2013, pp. 521--526.

\bibitem{tripicchio2015towards}
P.~Tripicchio, M.~Satler, G.~Dabisias, E.~Ruffaldi, and C.~A. Avizzano,
  ``Towards smart farming and sustainable agriculture with drones,'' in
  \emph{2015 International Conference on Intelligent Environments}.\hskip 1em
  plus 0.5em minus 0.4em\relax IEEE, 2015, pp. 140--143.

\bibitem{mazzia2020real}
V.~Mazzia, A.~Khaliq, F.~Salvetti, and M.~Chiaberge, ``Real-time apple
  detection system using embedded systems with hardware accelerators: An edge
  ai application,'' \emph{IEEE Access}, vol.~8, pp. 9102--9114, 2020.

\bibitem{kang2020real}
H.~Kang, H.~Zhou, X.~Wang, and C.~Chen, ``Real-time fruit recognition and
  grasping estimation for robotic apple harvesting,'' \emph{Sensors}, vol.~20,
  no.~19, p. 5670, 2020.

\bibitem{luo2018vision}
L.~Luo, Y.~Tang, Q.~Lu, X.~Chen, P.~Zhang, and X.~Zou, ``A vision methodology
  for harvesting robot to detect cutting points on peduncles of double
  overlapping grape clusters in a vineyard,'' \emph{Computers in Industry},
  vol.~99, pp. 130--139, 2018.

\bibitem{mohanty2016using}
S.~P. Mohanty, D.~P. Hughes, and M.~Salath{\'e}, ``Using deep learning for
  image-based plant disease detection,'' \emph{Frontiers in plant science},
  vol.~7, p. 1419, 2016.

\bibitem{ferentinos2018deep}
K.~P. Ferentinos, ``Deep learning models for plant disease detection and
  diagnosis,'' \emph{Computers and Electronics in Agriculture}, vol. 145, pp.
  311--318, 2018.

\bibitem{khaliq2019refining}
A.~Khaliq, V.~Mazzia, and M.~Chiaberge, ``Refining satellite imagery by using
  uav imagery for vineyard environment: A cnn based approach,'' in \emph{2019
  IEEE International Workshop on Metrology for Agriculture and Forestry
  (MetroAgriFor)}.\hskip 1em plus 0.5em minus 0.4em\relax IEEE, 2019, pp.
  25--29.

\bibitem{mazzia2020uav}
V.~Mazzia, L.~Comba, A.~Khaliq, M.~Chiaberge, and P.~Gay, ``Uav and machine
  learning based refinement of a satellite-driven vegetation index for
  precision agriculture,'' \emph{Sensors}, vol.~20, no.~9, p. 2530, 2020.

\bibitem{aghi2020autonomous}
D.~Aghi, V.~Mazzia, and M.~Chiaberge, ``Autonomous navigation in vineyards with
  deep learning at the edge,'' in \emph{International Conference on Robotics in
  Alpe-Adria Danube Region}.\hskip 1em plus 0.5em minus 0.4em\relax Springer,
  2020, pp. 479--486.

\bibitem{aghi2020local}
------, ``Local motion planner for autonomous navigation in vineyards with a
  rgb-d camera-based algorithm and deep learning synergy,'' \emph{Machines},
  vol.~8, no.~2, p.~27, 2020.

\bibitem{riggio2018low}
G.~Riggio, C.~Fantuzzi, and C.~Secchi, ``A low-cost navigation strategy for
  yield estimation in vineyards,'' in \emph{2018 IEEE International Conference
  on Robotics and Automation (ICRA)}.\hskip 1em plus 0.5em minus 0.4em\relax
  IEEE, 2018, pp. 2200--2205.

\bibitem{astolfi2018vineyard}
P.~Astolfi, A.~Gabrielli, L.~Bascetta, and M.~Matteucci, ``Vineyard autonomous
  navigation in the echord++ grape experiment,'' \emph{IFAC-PapersOnLine},
  vol.~51, no.~11, pp. 704--709, 2018.

\bibitem{barawid2007development}
O.~C. Barawid~Jr, A.~Mizushima, K.~Ishii, and N.~Noguchi, ``Development of an
  autonomous navigation system using a two-dimensional laser scanner in an
  orchard application,'' \emph{Biosystems Engineering}, vol.~96, no.~2, pp.
  139--149, 2007.

\bibitem{mazzia2019deep}
V.~Mazzia, A.~Tartaglia, M.~Chiaberge, and D.~Gandini, ``Deep learning
  algorithms for complex pattern recognition in ultrasonic sensors arrays,'' in
  \emph{International Conference on Machine Learning, Optimization, and Data
  Science}.\hskip 1em plus 0.5em minus 0.4em\relax Springer, 2019, pp. 24--35.

\bibitem{zoto2019automatic}
J.~Zoto, M.~A. Musci, A.~Khaliq, M.~Chiaberge, and I.~Aicardi, ``Automatic path
  planning for unmanned ground vehicle using uav imagery,'' in
  \emph{International Conference on Robotics in Alpe-Adria Danube
  Region}.\hskip 1em plus 0.5em minus 0.4em\relax Springer, 2019, pp. 223--230.

\bibitem{vidovic2014center}
I.~Vidovi{\'c} and R.~Scitovski, ``Center-based clustering for line detection
  and application to crop rows detection,'' \emph{Computers and electronics in
  agriculture}, vol. 109, pp. 212--220, 2014.

\bibitem{comba2018unsupervised}
L.~Comba, A.~Biglia, D.~R. Aimonino, and P.~Gay, ``Unsupervised detection of
  vineyards by 3d point-cloud uav photogrammetry for precision agriculture,''
  \emph{Computers and Electronics in Agriculture}, vol. 155, pp. 84--95, 2018.

\bibitem{misra2019mish}
D.~Misra, ``Mish: A self regularized non-monotonic neural activation
  function,'' \emph{arXiv preprint arXiv:1908.08681}, 2019.

\bibitem{woo2018cbam}
S.~Woo, J.~Park, J.-Y. Lee, and I.~So~Kweon, ``Cbam: Convolutional block
  attention module,'' in \emph{Proceedings of the European conference on
  computer vision (ECCV)}, 2018, pp. 3--19.

\bibitem{ronneberger2015u}
O.~Ronneberger, P.~Fischer, and T.~Brox, ``U-net: Convolutional networks for
  biomedical image segmentation,'' in \emph{International Conference on Medical
  image computing and computer-assisted intervention}.\hskip 1em plus 0.5em
  minus 0.4em\relax Springer, 2015, pp. 234--241.

\bibitem{redmon2016you}
J.~Redmon, S.~Divvala, R.~Girshick, and A.~Farhadi, ``You only look once:
  Unified, real-time object detection,'' in \emph{Proceedings of the IEEE
  conference on computer vision and pattern recognition}, 2016, pp. 779--788.

\bibitem{liu2016ssd}
W.~Liu, D.~Anguelov, D.~Erhan, C.~Szegedy, S.~Reed, C.-Y. Fu, and A.~C. Berg,
  ``Ssd: Single shot multibox detector,'' in \emph{European conference on
  computer vision}.\hskip 1em plus 0.5em minus 0.4em\relax Springer, 2016, pp.
  21--37.

\bibitem{ester1996density}
M.~Ester, H.-P. Kriegel, J.~Sander, X.~Xu \emph{et~al.}, ``A density-based
  algorithm for discovering clusters in large spatial databases with noise.''
  in \emph{Kdd}, vol.~96, no.~34, 1996, pp. 226--231.

\bibitem{matas2000robust}
J.~Matas, C.~Galambos, and J.~Kittler, ``Robust detection of lines using the
  progressive probabilistic hough transform,'' \emph{Computer vision and image
  understanding}, vol.~78, no.~1, pp. 119--137, 2000.

\bibitem{hart1968formal}
P.~E. Hart, N.~J. Nilsson, and B.~Raphael, ``A formal basis for the heuristic
  determination of minimum cost paths,'' \emph{IEEE transactions on Systems
  Science and Cybernetics}, vol.~4, no.~2, pp. 100--107, 1968.

\bibitem{Craw2017}
\BIBentryALTinterwordspacing
S.~Craw, \emph{Manhattan Distance}.\hskip 1em plus 0.5em minus 0.4em\relax
  Boston, MA: Springer US, 2017, pp. 790--791. [Online]. Available:
  \url{https://doi.org/10.1007/978-1-4899-7687-1_511}
\BIBentrySTDinterwordspacing

\bibitem{costa2019survey}
M.~M. Costa and M.~F. Silva, ``A survey on path planning algorithms for mobile
  robots,'' in \emph{2019 IEEE International Conference on Autonomous Robot
  Systems and Competitions (ICARSC)}.\hskip 1em plus 0.5em minus 0.4em\relax
  IEEE, 2019, pp. 1--7.

\bibitem{kingma2014adam}
D.~P. Kingma and J.~Ba, ``Adam: A method for stochastic optimization,''
  \emph{arXiv preprint arXiv:1412.6980}, 2014.

\bibitem{smith2017cyclical}
L.~N. Smith, ``Cyclical learning rates for training neural networks,'' in
  \emph{2017 IEEE Winter Conference on Applications of Computer Vision
  (WACV)}.\hskip 1em plus 0.5em minus 0.4em\relax IEEE, 2017, pp. 464--472.

\bibitem{abadi2016tensorflow}
M.~Abadi, P.~Barham, J.~Chen, Z.~Chen, A.~Davis, J.~Dean, M.~Devin,
  S.~Ghemawat, G.~Irving, M.~Isard \emph{et~al.}, ``Tensorflow: A system for
  large-scale machine learning,'' in \emph{12th $\{$USENIX$\}$ symposium on
  operating systems design and implementation ($\{$OSDI$\}$ 16)}, 2016, pp.
  265--283.

\bibitem{everingham2010pascal}
M.~Everingham, L.~Van~Gool, C.~K. Williams, J.~Winn, and A.~Zisserman, ``The
  pascal visual object classes (voc) challenge,'' \emph{International journal
  of computer vision}, vol.~88, no.~2, pp. 303--338, 2010.

\bibitem{lin2014microsoft}
T.-Y. Lin, M.~Maire, S.~Belongie, J.~Hays, P.~Perona, D.~Ramanan,
  P.~Doll{\'a}r, and C.~L. Zitnick, ``Microsoft coco: Common objects in
  context,'' in \emph{European conference on computer vision}.\hskip 1em plus
  0.5em minus 0.4em\relax Springer, 2014, pp. 740--755.

\end{thebibliography}

\vfill\break

\begin{IEEEbiography}[{\includegraphics[width=1in,height=1.25in,clip,keepaspectratio]{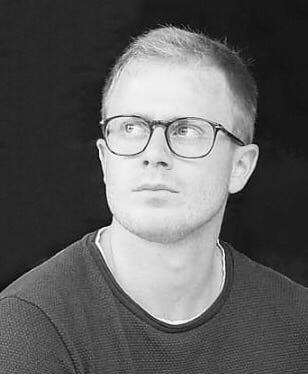}}]{Vittorio Mazzia} is a Ph.D. student in Electrical, Electronics and Communications Engineering working with the two Interdepartmental Centres PIC4SeR (\url{https://pic4ser.polito.it/}) and SmartData (\url{https://smartdata.polito.it/}). He received a master's degree in Mechatronics Engineering from the Politecnico di Torino, presenting a thesis entitled "Use of deep learning for automatic low-cost detection of cracks in tunnels," developed in collaboration with the California State University. His current research interests involve deep learning applied to different tasks of computer vision, autonomous navigation for service robotics, and reinforcement learning. Moreover, making use of neural compute devices (like Jetson Xavier, Jetson Nano, Movidius Neural Stick) for hardware acceleration, he is currently working on machine learning algorithms and their embedded implementation for AI at the edge. 
\end{IEEEbiography}

\begin{IEEEbiography}[{\includegraphics[width=1in,height=1.25in,clip,keepaspectratio]{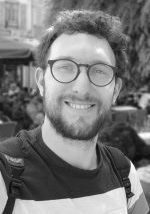}}]{Francesco Salvetti} is currently a Ph.D. student in Electrical, Electronics and Communications Engineering in collaboration with the two interdepartmental centers PIC4SeR (\url{https://pic4ser.polito.it/}) and Smart Data (\url{https://smartdata.polito.it/}) at Politecnico di Torino, Italy. He received his Bachelor's Degree in Electronic Engineering§ in 2017 and his Master’s Degree in Mechatronics Engineering in 2019 at Politecnico di Torino. He is currently working on Machine Learning applied to Computer Vision and Image Processing in robotics applications.
\end{IEEEbiography}

\begin{IEEEbiography}[{\includegraphics[width=1in,height=1.25in,clip,keepaspectratio]{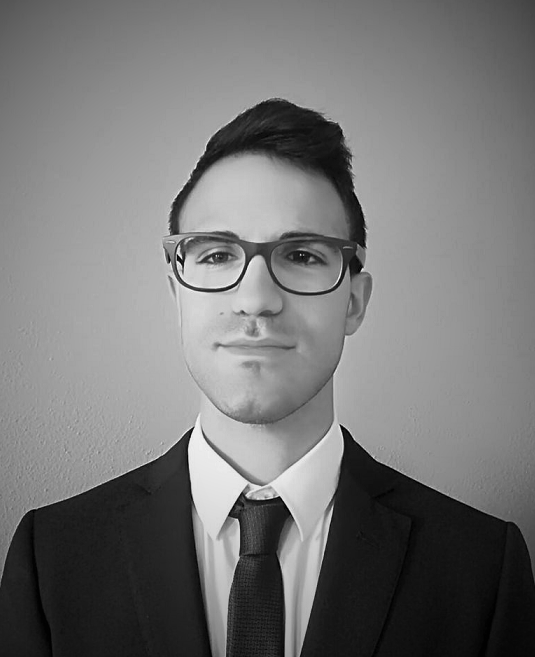}}]{Diego Aghi} is a researcher at PIC4SeR - PoliTO
Interdepartmental Centre for Service Robotics (\url{https://pic4ser.polito.it/}). He graduated from Politecnico di Torino with the thesis Navigation Algorithms for Unmanned Ground Vehicles in Precision Agriculture Applications carried out at PIC4SeR. He is now focusing his research activity on the development of machine learning and computer vision algorithms for autonomous navigation applications in the outdoor environment. 
\end{IEEEbiography}

\begin{IEEEbiography}[{\includegraphics[width=1in,height=1.25in,clip,keepaspectratio]{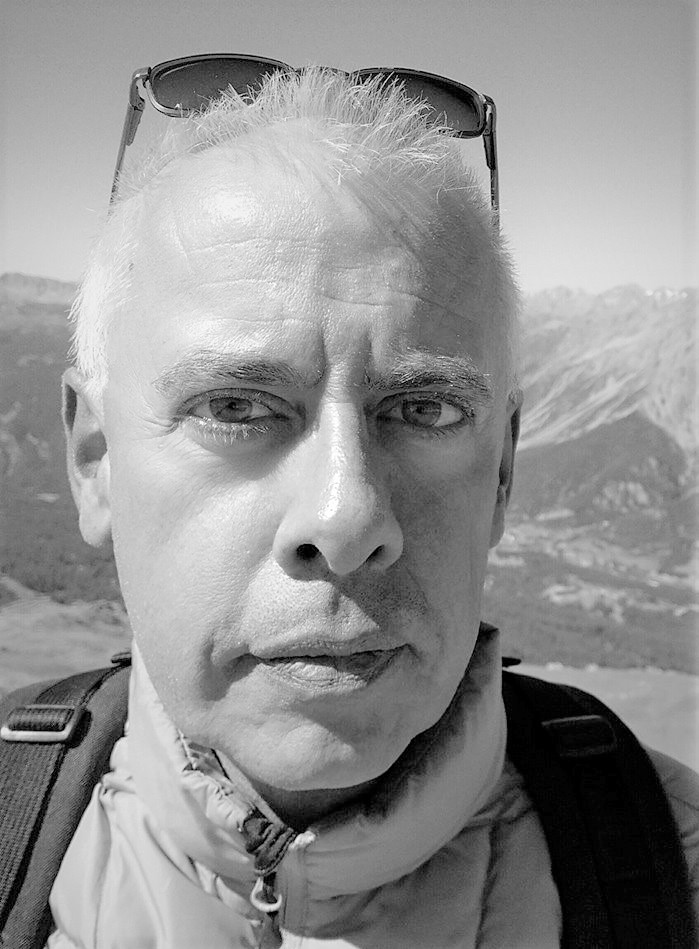}}]{Marcello Chiaberge} is currently an Associate Professor within the Department of Electronics and Telecommunications, Politecnico di Torino, Turin, Italy. He is also the Co-Director of the Mechatronics Lab, Politecnico di Torino
(\url{www.lim.polito.it}), Turin, and the Director and the Principal Investigator of the new Centre for Service Robotics (PIC4SeR, \url{https://pic4ser.polito.it/}), Turin. He has authored more than 100 articles accepted in international conferences and journals,
and he is the co-author of nine international patents. His research interests include
hardware implementation of neural networks and fuzzy systems and the design and implementation of reconfigurable real-time computing architectures.
\end{IEEEbiography}
\vfill

\end{document}